\documentclass{ar-1col}

\makeatletter%
\def\filter@val(#1)\@nil{\gdef\@currentlabel{#1}}
\def\exi#1{\item[#1]\expandafter\filter@val#1\@nil}
\makeatother

\usepackage{natbib}
\usepackage{url}
\usepackage{linguex}

\jname{Annual Reviews of Linguistics}
\jvol{AA}
\jyear{YYYY}
\doi{10.1146/10.1146/annurev-linguistics-032020-051035}

\newlength{\gap}
\begin{document}

\markboth{Linzen and Baroni}{Syntactic Structure from Deep Learning}

\title{Syntactic Structure from Deep Learning}

\author{Tal Linzen$^1$ and Marco Baroni$^{2,3,4}$
\affil{$^1$Department of Cognitive Science, John Hopkins University, Baltimore, US, MD 21218; email: tal.linzen@jhu.edu}
\affil{$^2$Facebook AI Research, Paris, France, 75002;  email: mbaroni@fb.com}
\affil{$^3$Catalan Institute for Research and Advanced Studies, Barcelona, Spain, 08010}
\affil{$^3$Departament de Traducci\'o i C\`encies del Llenguatge, Universitat Pompeu Fabra, Spain, 08018}}

\begin{abstract}
  Modern deep neural networks achieve impressive performance in engineering applications that require extensive linguistic skills, such as machine translation. This success has sparked interest in probing whether these models are inducing human-like grammatical knowledge from the raw data they are exposed to, and, consequently, whether they can shed new light on long-standing debates concerning the innate structure necessary for language acquisition. In this article, we survey representative studies of the syntactic abilities of deep networks, and discuss the broader implications that this work has for theoretical linguistics.
\end{abstract}

\maketitle

\tableofcontents

\section{INTRODUCTION}

In the last decade, artificial neural networks, rebranded as ``deep learning'' \citep{LeCun:etal:2015}, have made an astounding comeback in a range of technological applications. Among those applications are natural language processing (NLP) tasks ranging from machine translation \citep{Edunov:etal:2018} to reading comprehension \citep{Cui:etal:2017}. From a linguist's perspective, the applied success of deep neural networks (DNNs) is striking, because, unlike the systems that were popular in NLP a decade ago \citep{Jurafsky:Martin:2008}, DNNs' input data and architectures are not based on the symbolic representations familiar from linguistics, such as parse trees or logical formulas. Instead, DNNs learn to encode words and sentences as vectors (sequences of real numbers); these vectors, which do not bear a transparent relationship to classic linguistic structures, are then transformed through a series of simple arithmetic operations to produce the network's output. Any grammatical competence acquired by standard DNNs derives, then, from exposure to large amounts of raw text, combined with generic architectural features that have little to do with those that many linguists have deemed necessary for language acquisition, such as a preference for rules based on hierarchical tree structure \citep{Chomsky:1986}.

Ostensibly, the success of deep learning invites a reassessment of classic arguments that language acquisition necessitates rich innate structure. But measures of practical success do not directly engage with the evidence that has motivated structural assumptions in linguistics: whereas success in applications rests primarily on the system's ability to handle common constructions, it is often the rare constructions that are the most informative ones from a theoretical standpoint. In this survey, we focus on work that directly evaluates DNNs' syntactic knowledge using paradigms from linguistics and psycholinguistics that highlight such theoretically significant cases. After a brief introduction to deep learning for language processing (Section \ref{sec:deep-learning}), we review work applying (psycho)linguistic analysis methods to English subject-verb number agreement, filler-gap dependencies, and other syntactic phenomena (Sections \ref{sec:long-distance-agreement} and \ref{sec:other}); this body of work suggests that contemporary DNNs can learn a surprising amount about syntax, but fall short of human competence. We then briefly survey work that aims to illuminate the internal processes by which DNNs accomplish their grammatical behavior (Section~\ref{sec:internal-representations}). Finally, we discuss our view of the implications of this body of work for linguistic theory (Section~\ref{sec:discussion}).

\section{DEEP LEARNING FOR LANGUAGE PROCESSING}
\label{sec:deep-learning}

This section provides a very short overview of artificial neural networks as applied to language processing at the word and sentence level; for a contemporary book-length introduction to neural networks for language processing, see \citet{Goldberg:2017}; for a less technical introduction to artificial neural networks for cognitive scientists, which does not include a treatment of language processing, see \citet{elman1998rethinking}.  

    Artificial neural networks are mathematical objects that compute functions from one sequence of real numbers to another sequence. They do so using large collections of simple computation units (``neurons''). Each of these units calculates a weighted average of its inputs; this weighted average is then passed as an input to a simple nonlinear function, such as the sigmoid ($\sigma(a)=1/(1+e^{-a})$). In other words, the function computed by each unit is $\sigma(w_1x_1+\ldots+w_nx_n)$, where $w_1,\ldots,w_n$ are the weights, and $x_1,\ldots,x_n$ are the inputs. Although the computation performed by each unit is very simple, by arraying a large number of units in layers, such that the output of the units in one layers serves as the input to the units in the following layer, much more complex functions can be computed (theoretically, all functions can be approximated to arbitrary precision; \citealt{Leshno:etal:1993}). The presence of multiple layers, incidentally, is what makes the networks ``deep''.

    The network's weights are not set by the designer of the system, but are learned from examples. Each such ``training'' example consists of an input $\mathbf{x}_i$ and an expected output~$y_i$. The training procedure starts from a random set of weights. The network then iterates through each input example and computes the output $\hat{y}_i$ based on the current weights. This output is then compared to the expected output $y_i$, and the weights are adjusted by a small amount such that the next time the DNN receives $\mathbf{x}_i$ as its input, the discrepancy between $\hat{y}_i$ and $y_i$ will be smaller (this process is referred to as gradient descent).

Neural networks compute numerical functions. In order for them to process language, each input and output word needs to be encoded as a vector (a sequence of real numbers). In principle, a word could be encoded by a vector whose size is equal to the size of the vocabulary, and which has zeros everywhere except for one component that indicates the identity of the word (a ``one-hot vector''); in practice, however, word are typically encoded by vectors that are much smaller and denser (have nonzero values in most components). Such distributed representations, or word embeddings, make it possible to assign similar vectors to words that occur in similar contexts or have similar meanings (e.g., $(1, 2.5, 3)$ for \emph{table} and $(1.2, 2.5, 2.8)$ for \emph{desk}, but $(2.1, -3, 4)$ for \emph{dog}). Rather than being set by the designer of the system, these word embeddings are learned using gradient descent, just like the network's weights.

    Word embeddings provide a mechanism for encoding individual words, but additional machinery is needed to process sequences of words. One way to do this is using recurrent neural networks (RNNs). An RNN processes the sentence from left to right, maintaining a single vector $\mathbf{h}_t$, the ``hidden state'', which represents the first $t$ words of the sentence. The next hidden state, $\mathbf{h}_{t+1}$, is computed from $\mathbf{h}_t$ and the embedding of the $t+1$-th word using standard neural network arithmetic. The hidden state thus acts as a bottleneck: the network does not have access to its earlier hidden states $\mathbf{h}_1,\ldots,\mathbf{h}_{t-1}$ when computing $\mathbf{h}_{t+1}$. 

    The performance of RNNs can be improved by the addition of gates. Gated RNNs, such as Long Short-Term Memory networks (LSTMs, \citealt{Hochreiter:Schmidhuber:1997}) and gated recurrent units (GRUs, \citealt{Cho:etal:2014}), possess a mechanism that allows them to better control the extent to which information in their hidden state is updated after each word. At least in principle, gating enables the network to better track dependencies that span a large number of words. LSTMs have become the \emph{de-facto} standard RNN variant in NLP, and most of the studies we review here use this architecture.

    Attention, another important innovation, relaxes the single-hidden-state bottleneck, allowing the network to take all of its previous hidden states into account when computing the next state \citep{Bahdanau:etal:2015}. As is the case for other DNN architectural elements, the dynamics of both gating and attention are not hard-coded, but rather controlled by weights learned during the training phase.  As it turns out, when the network has access to a large window of previous states through attention, the recurrence mechanism that privileges the computation of $\mathbf{h}_{t+1}$ from $\mathbf{h}_t$ alone becomes largely redundant. Consequently, some state-of-the-art sequence processing architectures, such as the transformer \citep{Vaswani:etal:2017}, dispense with recurrence altogether and rely on attention only to carry information across time.  Using specialized hardware, transformers can be trained effectively on very large corpora, and they have become very common in NLP systems such as BERT \citep{Devlin:etal:2019}; however, a clear picture of the differences between their linguistic abilities and those of RNNs, especially when both are exposed to the same amount of training data, has yet to emerge \citep{tran2018importance,goldberg2019assessing,rogers2020primer}.  

DNNs for language processing are used in three common settings. When used as a classifier, the network outputs a discrete label for the sequence; for example, a binary acceptability judgment system might output either ``acceptable'' or ``unacceptable''. This is a supervised setup, where the network is trained on a corpus of example sentences annotated for acceptability. 

When used as a ``language model'', the network receives the first $n$ words of a sentence, and assigns a probability to each of the words that could come up next. The network is trained to maximize the probability of the word that actually occurs next in the sentence, as it appears in the corpus. For example, given the first five words of sentence \emph{the children went outside to play}, the training objective would be to assign the largest probability possible to the final word \emph{play}.  This setup is unsupervised, in the sense that all we need to train the network is a text corpus, without any annotation.  

Finally, in the sequence-to-sequence (``seq2seq'') setting, the network is expected to generate an output sequence in response to an input sequence, for example when translating between languages \citep{Sutskever:etal:2014}. This is done by chaining two RNNs, where the last state of the encoder RNN serves as input to the decoder RNN. Attention can be used to allow the decoder to base its decisions on all of the encoders' hidden states rather than the last one only. The seq2seq setting requires pairs of input and output sequences as training materials (e.g., sentences in a source language, and the corresponding translations in a target language), but do not require any further annotation.

\section{LONG-DISTANCE AGREEMENT}
\label{sec:long-distance-agreement}

Much of the initial analysis work on the syntactic abilities of DNNs centered on long-distance agreement between subject and verb, or between other elements in a syntactic dependency. Subject-verb agreement is a paradigmatic example of the observation that words in sentences are organized according to ``structures, not strings'' \citep{Everaert:etal:2015}: the notion of the subject makes crucial reference to the abstract structure of the sentence, rather than to the linear sequence of the words it is made up of. A DNN's ability to capture subject-verb agreement can be evaluated in a straightforward manner using the \emph{number prediction task} \citep{Linzen:etal:2016}. In this setting, the DNN is exposed to a sentence prefix such as Example \ref{na}, where the word following the prefix is expected to be a verb; the DNN is then tasked with predicting whether the verb that is expected to come up next should be in plural or singular form:

\ex. The \textbf{length} of the \emph{forewings} (is/*are)\ldots \label{na}

The correct form depends on the number of the verb's subject, which, in Example \ref{na}, is determined by its head \emph{length}, rather than by \emph{forewings}. Nouns such as \emph{forewings}, which intervene between the head of the subject and the verb, are referred to as attractors \citep{Bock:Miller:1991}. To correctly predict the number of the verb, then, the DNN must derive an implicit analysis of the structure of the sentence, and resist the lure of the proximal but irrelevant attractor.

Linzen and colleagues exposed an LSTM---one of the most popular DNN architectures for sentence processing (Section \ref{sec:deep-learning})---to a set of corpus-extracted English sentence prefixes, such as the one in Example \ref{na}, and trained it to perform the number prediction task (that is, it was trained in the supervised classification setting, in the terms of Section~\ref{sec:deep-learning}). When tested on new prefixes that were not presented to it during training, the network correctly predicted the number of the upcoming verb more than $99\%$ of the time.  However, since in the overwhelming majority of English sentences the head of the subject happens to be the most recent noun preceding the verb (as it is, for example, in \textit{the well-known lawyers (are)}), overall accuracy is not directly informative about whether the DNN is able to identify the head of the subject. Crucially, even when it was tested on sentences with attractors, the DNN showed considerable robustness: in sentences with as many as four attractors, number prediction accuracy was still 82\%. While the probability of an error increased substantially in the presence of attractors, then, it was considerably lower than what would be expected if the network was typically misled by attractors.  

Extending these results, \cite{Bernardy:Lappin:2017} showed that other DNN architectures (GRUs and convolutional networks) can also tackle the number prediction task with considerable success. This indicates that Linzen's original result does not crucially depend on specific features of the LSTM architecture. In particular, the convolutional network trained by \citeauthor{Bernardy:Lappin:2017} completely dispenses with the recurrent mechanism central to LSTMs \citep{Kalchbrenner:etal:2014}. Its success therefore points to the generality of the outcome across current DNN
models.

In these early studies, the DNNs were trained specifically to predict the number of an upcoming verb, and were given explicit feedback about verb number in a large set of sentences.  \cite{Gulordava:etal:2018} showed that LSTMs can learn a significant amount about long-distance agreement even when trained simply to predict the next word in a corpus, without any specific focus on subject-verb agreement or number features (the language modeling setting).  To test the trained DNN's ability to compute agreement, the authors exposed it to a sentence prefix, and compared the probabilities it assigned, given the prefix, to the singular and plural forms of the upcoming verb. The DNN was considered to have made a correct number prediction if it assigned a higher probability to the contextually appropriate form; after \textit{the length of the forewings}, for example, the probability of \textit{is} was expected to be higher than that of \textit{are}.

Using this methodology, Gulordava and colleagues showed that LSTMs trained only to predict the next word show high agreement prediction accuracy when tested on sentence prefixes extracted from Wikipedia, across four languages (English, Hebrew, Italian and Russian), and in dependencies beyond subject-verb agreement, in languages that have them (e.g., adjective-noun agreement). When compared to human subjects, the LSTM's agreement prediction accuracy in Italian (the only language for which this comparison was carried out) was only moderately lower. Finally, Gulordava and colleagues tested the DNN's predictions in ``colorless green ideas'' prefixes: grammatically well-formed but highly semantically implausible prefixes, constructed by replacing the content words in prefixes from the corpus with other words from the same syntactic category (e.g., \emph{the colorless green ideas near the duck (are/*is)\ldots}). The DNN showed only a mild degradation in performance on these sentences, suggesting that it is capable of computing agreement in the absence of lexical or semantic cues. Overall, this study suggests that training on word prediction alone, without additional syntactic supervision or semantic grounding, can teach networks a substantial amount about long-distance agreement dependencies and the syntactic categories that underlie them (such as the notion of a syntactic subject).

This is not to say that LSTM language models acquire perfect syntactic competence. There is evidence that they rely on simple heuristics. For example, because English does not indicate the end of a relative clause with an explicit marker, they tend to expect embedded clauses to be relatively short \citep{Linzen:Leonard:2018}; and they pay undue attention to the number of the first noun of the sentence, even when it is not the head of the subject \citep{Kuncoro:etal:2018b}. In a study using controlled experimental materials, instead of evaluation sentences sampled from a corpus, \citet{Marvin:Linzen:2018} found that LSTMs performed poorly on some sentence types that are infrequent in corpora, such as nested agreement dependencies across an object relative clause (\textit{The \textit{farmer} that the parents love (swims/*swim)}). At the same time, the fact that DNNs perform consistently well across a range of agreement dependencies suggests that they are able to extract certain abstract syntactic generalizations. In Section~\ref{sec:internal-representations} below, we will discuss work that brings us close to a mechanistic understanding of how LSTMs perform long-distance agreement.

\section{OTHER SYNTACTIC PHENOMENA}
\label{sec:other}

Research on the syntactic abilities of neural networks quickly expanded beyond agreement to include a variety of other syntactic phenomena. Here, we review a few representative lines of work. For additional examples, we refer the reader to the proceedings of the BlackBox NLP workshops \citep{Linzen:etal:2018,Linzen:etal:2019} and of the Society for Computation in Linguistics.

\cite{Wilcox:etal:2018} tested the sensitivity of LSTM language models to English filler-gap dependencies. As in \citet{Gulordava:etal:2018}, the DNNs were trained only to predict the next word, without any specific supervision on this construction. In a filler-gap dependency, a wh-licensor sets up a prediction for a gap---one of the noun phrases in the embedded clause must be omitted:

\ex.\label{gazelle1}\a.I know that you insulted your aunt yesterday. \textit{(no wh-licensor, no gap)}\label{nowh_nogap} 
\b. *I know who you insulted your aunt yesterday. \textit{(wh-licensor, no gap)}\label{wh_nogap}

According to Wilcox and colleagues' logic, if the network is sensitive to this constraint, we expect it to be more surprised by  \textit{yesterday} in the ungrammatical Example \ref{wh_nogap} than in the grammatical \ref{nowh_nogap} (we say that the DNN is more surprised by a word if it assigns a lower probability to that word). The DNN's surprise is measured at \textit{yesterday}---instead of at the filled gap \textit{your aunt}, which is arguably the locus of ungrammaticality---because there are contexts in which the sentence can be continued in a grammatical way after this noun phrase (\textit{I know who you insulted your aunt with \underline{\hspace{\gap}}}). 

It is not only the case that gaps are required after a wh-licensor; they are \textit{only allowed} in the presence of such a licensor. If the DNN is fully sensitive to filler-gap dependencies, then, we expect it to be more surprised by \textit{yesterday} in Example \ref{nowh_gap}, where the gap is ungrammatical, than in \ref{wh_gap}:

\ex.\label{gazelle2}\a. *I know that you insulted \underline{\hspace{\gap}} yesterday.  \textit{(no wh-licensor, gap)}\label{nowh_gap}
\b. I know who you insulted \underline{\hspace{\gap}} yesterday. \textit{(wh-licensor, gap)}\label{wh_gap}

\citet{Wilcox:etal:2018} report that the networks showed the expected pattern. This was the case not only for direct object extraction as in Examples \ref{gazelle1} and \ref{gazelle2}, but also for subject extraction (Example \ref{subj-extraction}) and indirect object extraction (Example \ref{prep-extraction}):

\ex.\label{filler-gap}
\a.I know who \underline{\hspace{\gap}} showed the presentation to the visitors yesterday.\label{subj-extraction}      
\b.I know who the businessman showed the presentation to \underline{\hspace{\gap}} yesterday.\label{prep-extraction} 

The acceptability of grammatical filler-gap constructions was only
marginally affected by the distance between the wh-licensor and
the gap, in line with studies of human processing. Further, the networks correctly learned that a wh-phrase cannot license more than one gap, distinguishing the well-formed Examples \ref{single-gap-obj} and \ref{single-gap-subj} from the less natural \ref{double-gap}:

\ex.\a.I know what the lion devoured  \underline{\hspace{\gap}} at sunrise.\label{single-gap-obj}
\b.I know what  \underline{\hspace{\gap}} devoured a mouse at sunrise.\label{single-gap-subj}
\c.*I know what \underline{\hspace{\gap}} devoured  \underline{\hspace{\gap}} at sunrise.\label{double-gap}

In some syntactic configurations, referred to as ``islands'' \citep{Ross:1967}, extracting a noun phrase (replacing it with a gap) is ungrammatical. If the networks are sensitive to this constraint, their expectation of a gap should be attenuated in these contexts. Wilcox and colleagues found this to be the case for some contexts, such as wh-islands and adjunct islands. One exception was the Subject Island Constraint, under which a prepositional phrase modifying a noun phrase can only contain a gap if that noun phrase is not the subject:

\ex.\a.I know who the family bought the painting by \underline{\hspace{\gap}} last year.\label{pp-extraction-obj}      \b.*I know who the painting by \underline{\hspace{\gap}} fetched a high price at auction.\label{pp-extraction-subj}      

Neither of the two LSTMs tested by Wilcox captured this asymmetry. Interestingly, despite the fact that the two LSTMs were quite similar in terms of architecture and training corpus, they erred in opposite ways, with one finding both Examples \ref{pp-extraction-obj} and \ref{pp-extraction-subj} grammatical, and the other both unacceptable.

Overall, Wilcox and colleagues' conclusions about LSTM language models' sensitivity to filler-gap dependencies are quite upbeat. Other authors have reached more mixed conclusions. \citet{Chowdhury:Zamparelli:2018} argue that what appears to be an effect of syntactic islands on language model probabilities can be explained using other, non-grammatical factors. \citet{chaves2020fillergap} shows that DNNs do not capture the full complexity of island constraints, for example in negative islands (such as \textit{*How fast didn't John drive \underline{\hspace{\gap}}?}), where semantics and pragmatics play a central role. And \citet{warstadt2019blimp} report that DNNs that displayed significant sensitivity to a range of syntactic phenomena showed limited sensitivity to the island constraints they tested. In future research, it would be fruitful to establish how these mixed results arise from differences in the particular constructions and evaluation measures used in these studies.

\citet{Futrell:etal:2019} provide converging evidence that LSTMs can keep track of syntactic state. In a similar paradigm to that of \citet{Wilcox:etal:2018}, the LSTM language model they tested showed high surprise when a subordinate clause was not followed by a matrix clause, as in Example~\ref{subordinate_notmatrix}:

\ex.*As the doctor studied the textbook. \label{subordinate_notmatrix}

In so-called ``NP/Z'' garden path sentences such as Example~\ref{garden_path}, in which a noun phrase (\textit{the vet...}) that is preferentially attached as the direct object of the subordinate clause verb (\textit{scratched}) later turns out to be the subject of the matrix clause, the DNN was surprised at \textit{took}, as is the case with human subjects (see also \citealt{vanschijndel2018gardenpath}):

\ex.When the dog scratched the vet with his new assistant took off the muzzle.\label{garden_path}

Futrell and colleagues compared the LSTM of Gulordava and colleagues to two models trained on orders-of-magnitude less data: a standard LSTM, and a Recurrent Neural Network Grammar (RNNG; \citealt{Dyer:etal:2016}), which processes words based on the correct parse of the sentences, thus incorporating a strong bias towards syntactic structure. In this minimal-data regime, only the syntactically-biased RNNG was able to track syntactic state appropriately; however, the RNNG did not perform better than the Gulordava LSTM (which was trained on considerably more data).

Other studies have further investigated the ways in which the syntactic performance of a DNN is affected by its architecture and the amount of syntactic supervision it receives. \citet{McCoy:etal:2020} explored this question using the test case of auxiliary fronting in English question formation, a hierarchic phenomenon which has become the ``parade case'' of the poverty-of-the-stimulus argument \citep{Chomsky:1986}. Using the sequence-to-sequence framework (Section \ref{sec:deep-learning}), McCoy and colleagues trained a range of DNNs to produce questions from simple declarative sentences in a small fragment of English, as in the following pair:

\ex.
  \a.\label{question-formation-dec}
  The zebra \textbf{does} chuckle.
  \b.\label{question-formation-simple-quest}
  \textbf{Does} the zebra chuckle?

Following training, the DNNs were asked to generate questions from statements in which the subject was modified by a relative clause, as in Example \ref{question-formation-dec-rc}. This syntactic configuration, which was withheld from the DNNs' training data (following Chomsky's assumption that such configurations are very rare in child-directed speech), differentiates two generalizations that a learner could plausibly acquire. If the DNN learned, correctly, to front the main clause auxiliary (the \textsc{move-main} rule), it will produce Question~\ref{question-formation-move-main}. But the examples seen during training are also compatible with the non-hierarchical rule \textsc{move-first}, whereby the first auxiliary in the sentence is fronted irrespective of its syntactic role, as in Example~\ref{question-formation-move-first}.  

\ex.\a.\label{question-formation-dec-rc}
  Your zebras that \emph{don't} dance \textbf{do} chuckle.
  \b.\label{question-formation-move-main}
  \textbf{Do} your zebras that \emph{don't} dance chuckle?
  \c.\label{question-formation-move-first}
  \emph{Don't} your zebras that dance \textbf{do} chuckle?

Since the DNNs were trained only on examples that are ambiguous between the two rules, any preference for \textsc{move-main} would arise from the DNN's \emph{a priori} bias, and possibly a specific bias favoring hierarchical rules. 
And indeed, at least some of the DNN architectures McCoy and colleagues tested were biased in favor of \textsc{move-main}. But this bias was not very robust: small differences in network parameters, or even in random weight initializations of the exact same architecture, had a large effect on the outcome. Some differences are amenable to explanation: for example, gating mechanisms that disfavour counting lead to a preference for \textsc{move-main}, possibly because it is difficult to implement \textsc{move-first} without some counting device. Other factors are more difficult to interpret, especially because they interact in surprising ways: for example, different kinds of attention lead to more or less pronounced hierarchical behavior depending on the underlying gating mechanism. Finally, even the architectures that did acquired \textsc{move-main} for auxiliary fronting preferred a linear rule in a similarly ambiguous subject-verb agreement task, suggesting that these DNNs' bias is not reliably hierarchical.

McCoy and colleagues contrast the mixed results for standard sequential RNNs with the robust results of experiments with tree-based RNNs, which, like the RNNG architecture we briefly discussed above, combine the words of the sentence in an order determined by an explicit syntactic parse, rather than from left to right as in standard RNNs \citep{pollack1990recursive,socher2011parsing}. Such tree-based RNNs showed the clearest across-the-board preference for \textsc{move-main} (and for the hierarchical generalization in the agreement test). However, this robust preference emerged only when the bias was implemented in its strongest form: both the encoder and the decoder were based on trees, and correct trees were provided for both the input and the output. An important direction for future work is to determine whether reliably hierarchical generalization can arise from weaker architectural features.

Most studies of agreement and related phenomena have focused on languages with similar syntactic properties to English.  An interesting exception is provided by \citet{Ravfogel:etal:2018}, who studied case assignment in Basque. Basque has a number of characteristics that make it very different from English, such as relatively free word order, an ergative case system, and explicit marking of all arguments through morphemes suffixed to the verb form. Ravfogel and colleagues explored the task of reconstructing the case suffix (or lack thereof) of each word in sentences with all suffixes stripped off, e.g., reconstructing Example \ref{basque-example-full} from \ref{basque-example-no-suff}: 

\ex.
\a.\label{basque-example-full}      
\gll Kutxazain-ek bezeroa-ri liburu-ak eman dizkiote\\
cashier-\textsc{pl.erg} customer-\textsc{sg.dat} book-\textsc{pl.abs} gave they-them-to-her/him\\
\trans `The cashiers gave the books to the customer'
\b.\label{basque-example-no-suff}      
Kutxazaina bezeroa liburua eman dizkiote.

They formulated this as a supervised classification problem, which they tackled with a bi-directional LSTM, that is, an LSTM trained, in parallel, in both the left-to-right and right-to-left directions, in order to handle the fact that arguments can both precede and follow the verb. They achieved relatively high overall accuracy, with difficulties concentrating in particular around dative suffix prediction. A \emph{post-hoc} analysis revealed that the model was using a mix of shallow heuristics (e.g., relying on the closest verb to each argument to determine its case) and genuine generalizations (e.g., correctly encoding the distinction between the ergative and absolutive cases).

\section{WHAT DO THE NETWORK'S INTERNAL REPRESENTATIONS ENCODE?}
\label{sec:internal-representations}

The work reviewed so far evaluates a neural network's syntactic abilities by examining its output in response to inputs with particular syntactic properties. Much as linguists study human syntactic knowledge by providing or eliciting acceptability judgments, without access to internal brain states, this so-called ``black box'' approach does not require access to the network's inner workings. Compared to the neuroscientific techniques that are currently available for studying human subjects, however, direct access to the state of an artificial network is much cheaper, and it can be more granular: we know precisely what the activation of each unit is after the DNN processes each word. To the extent that the network's behavior indicates that it captures a particular syntactic phenomenon, then, there is clear interest in understanding how this behavior arises from the network's internal states.

This sort of analysis faces major challenges. In traditional symbolic systems, internal representations are in a format that is readily interpretable by a researcher (e.g., a parse tree or a logical formula), and so are the processes that operate over those representations (e.g., syntactic transformations or deduction rules). By contrast, the internal states of a DNN consist of vectors of hundreds or thousands of real numbers, and processing in the network involves applying to those vectors arithmetic operations that are parameterized by millions of weights. To understand how the DNN's behavior arises from its internal states, then, we need methods that allow us to translate these vectors into a format interpretable by a human. Such translation is difficult and may not be possible in all cases.  In this section, we review a handful of methods that attempt to interpret DNN internal states and link them to the network's behavior; for additional pointers, see \citet{belinkov2019analysis}.  

Perhaps the most popular interpretation method is based on diagnostic classifiers (or ``probing tasks''; \citealt{Shi:etal:2016,Adi:etal:2017}). This approach takes vectors produced by an existing network $N$---trained, for example, to represent sentences---and measures to what extent a new, separate classifier $C$ (which could be a simple neural network trained from scratch) can recover a particular linguistic distinction from $N$'s vector representations. This is done by presenting $C$ with $N$'s representations of sentences of different types---for example, sentences with a singular subject, and sentences with a plural subject, labeled as such---and training it to distinguish the two classes. If $C$ generalizes this distinction with high accuracy to the vector representations of new sentences, the conclusion is that a component of $N$ that received this vector representation as input could, in principle, have access to this information (this is analogous to multivoxel pattern analysis in neuroscience; \citealt{Haxby:etal:2001}).  

In a study of the mechanisms tracking subject-verb agreement in a DNN language model, for example, \citet{Giulianelli:etal:2018} showed that the plurality of the subject could be decoded with high accuracy from the hidden state of the DNN. In another study, \citet{Conneau:etal:2018} showed that a classifier could be trained to decode from a DNN's vector encoding of a sentence such syntactic information as the maximal depth of the parse tree of the sentence.  A related but distinct method is the structural probe of \citet{Hewitt:Manning:2019}, which seeks to find a simple similarity metric between a DNN's internal representations of words in context, such that the similarity between the vector representations of two words corresponds to the distance between the two words in a syntactic parse of the sentence. \citeauthor{Hewitt:Manning:2019} showed that such a similarity metric can be found for the representations generated by modern DNNs based on LSTMs or transformers, but not for simpler (baseline) representations.

An important caveat of the methods described so far is that successful recovery of information from the network's representation does not establish that a network in fact \textit{uses} that information: the information may not affect the network's behavior in any way. To illustrate this issue, in their study of auxiliary fronting, \citet{McCoy:etal:2018} analyzed the sentence  representations produced by various networks, by training two diagnostic classifiers: one that decodes the main verb's auxiliary, which is relevant to correct generalization on the auxiliary fronting task, and one that decodes the irrelevant first auxiliary (see Section~\ref{sec:other}). Disconcertingly for the naive interpretation of diagnostic classifier accuracy, both types of information were decodable with high accuracy from the networks, regardless of whether the DNNs' behavior indicated that they relied on one or the other. \citet{Giulianelli:etal:2018} addressed this concern in their diagnostic classifier study of subject-verb agreement, in two ways.  First, they showed that the classifier's accuracy was much lower in sentences in which the language model made incorrect agreement predictions; and second, they were able to use the classifier to \textit{intervene} in the state of the network and cause it to change its agreement predictions. Very few of the studies that use the diagnostic classifier method provide such compelling evidence linking the information decoded by the classifier to the analyzed network's behavior.  

The studies discussed so far have examined the numerical activation of a large set of the units of a DNN, jointly treated as a vector. \cite{Lakretz:etal:2019} explored whether the activation of an \textit{individual unit} had a causal effect on the behavior of the DNN they studied. They ablated each unit---that is, set its activation to zero---and measured how this affected the long-distance agreement performance of the \citet{Gulordava:etal:2018} LSTM language model (Section~\ref{sec:long-distance-agreement}). Using this method, Lakretz and colleagues uncovered a sparse mechanism whereby two units keep track of singular and plural number for agreement purposes. These units are in turn linked to a distributed circuit (that is, a circuit consisting of a number of units) that records the syntactic structure of a sentence, signaling when number information needs to be stored and released. This sophisticated grammar-aware sub-network is complemented by a bank of syntax-insensitive cells that apply agreement heuristics based on linear distance rather than syntactic structure.  

In sum, Lakretz and colleagues showed that the neurons of a LSTM trained to predict the next word implement a genuine syntax-based rule to track agreement. However, the sparse mechanism the DNN develops, complemented by the heuristic linear-distance system, cannot handle multiply embedded levels of long-distance relations. Interestingly, what becomes increasingly difficult for the network is not the outermost agreement, which is handled by the sparse circuit, but the embedded one: with the sparse circuit being occupied by outermost agreement tracking, agreement in the embedded clause can only rely on syntactically naive linear-distance units, which are fooled by intervening attractors. In other words, in a sentence such as Example \ref{multiple-embeddings}, the network has greater difficulty predicting the correct number for \emph{like} than \emph{is}.: 
\ex.\label{multiple-embeddings}
 The \textbf{kid}$_1$ that the \textbf{parents}$_2$ of our \emph{neighbour} \textbf{like}$_2$ \textbf{is}$_1$ tall.

Such a precise understanding of how agreement checking is (imperfectly) implemented in a DNN may help to formulate predictions about the processing of syntactic dependencies by humans \citep{lakretz2020limits}.

\section{DISCUSSION}
\label{sec:discussion}

In the past decade, deep learning has underpinned significant advances in NLP applications. The quality of deep-learning-based machine translation systems such as Google Translate and DeepL is sufficiently high that they have become useful in everyday life. Perhaps even more strikingly, DNNs trained only on large amounts of natural text---without ``innate'' linguistic constraints, and without support from explicit linguistic annotation---have shown an ability to generate long stretches of text that is grammatically valid, semantically consistent, and displays coherent discourse structure \citep{Radford:etal:2019}.  Here, we discuss to what extent this success should inform classic debates in linguistics about the innate mechanisms necessary for language acquisition, and about human linguistic competence more generally (see \citealt{Cichy:Kaiser:2019} for a high-level perspective on the place of DNNs in cognitive science).
\\
\\
\noindent{}\textbf{Nature vs.~nurture.} 
In early debates about the neural network approach to cognitive science (connectionism), neural networks were often portrayed as pure ``empiricist machines'', that is, as learners devoid of innate biases, which induce all their cognitive abilities from data \citep[e.g.,][]{Churchland:1989,Clark:1989,Fodor:Pylyshyn:1988,Pinker:Prince:1988,Christiansen:Chater:1999b}. If that is the case, DNNs' success on syntactic tasks may be taken to indicate that human-like syntactic competence can be acquired through mere ``statistical learning'', refuting the classic poverty-of-the-stimulus argument \citep{lasnik2017poverty}. But learning theory considerations show that the notion of a \emph{tabula rasa} is incoherent in practice. Finite training data is always consistent with an infinite number of possible generalizations: the stimulus is always poor. Consequently, any useful learner must have innate biases that would lead it to prefer some possible generalizations over others \citep{mitchell1980theneed}. DNNs are not an exception: they have biases that arise from their initial weights and from the structure of their architectures, which incorporate assumptions of temporal invariance, gating mechanisms, attention, encoding and decoding modules, etc. (see Section \ref{sec:deep-learning}). 

While DNNs are clearly not \emph{tabulae rasae}, their biases are quite different from those traditionally proposed by linguists as underlying language acquisition. DNNs used for NLP are not constrained to perform only syntactically defined, ``recursive'' operations, as dictated by the structure-sensitivity-of-rules principle \citep{chomsky1965,chomsky1980} or Merge \citep{Chomsky:1995,Hauser:etal:2002}. If anything, the central architectural features of standard DNNs emphasize sequential left-to-right processing (RNNs) and content-addressable memory storage and retrieval (gating, attention). If a DNN performs syntactic tasks in a way that is consistent with human syntactic competence, we can conclude, for example, that an innate principle constraining the system to use Merge is not needed to acquire the relevant abilities. What we cannot do is ignore the biases contributed by a DNN's architecture, and conclude that statistical learning from data alone suffices to acquire the relevant abilities. At the moment, we do not know which (if any) DNN architectural features are fundamental for learning syntax. Future work should tease apart the crucial factors that enable particular DNN architectures to generalize in a human-like way from finite training data, along the lines of studies such as \citet{McCoy:etal:2020}, which explicitly link success in diagnostic tasks to the specific priors of different DNN architectures, and \citet{Lakretz:etal:2019}, which characterize at a mechanistic level how architectural features such as gates underlie the ability of a DNN to perform a particular linguistic task.
\\
\\
\noindent{}\textbf{Incorporating linguistic principles into DNNs.} 
Instead of trying to understand how the somewhat opaque prior biases of DNNs affect their linguistic abilities, we can attempt to directly inject a particular bias of theoretical interest into their architecture, and assess the impact of that bias on the generalizations acquired by the DNN. The studies of \citet{Futrell:etal:2019} and \citet{McCoy:etal:2020} reviewed in Section \ref{sec:other} illustrate this approach. They showed that when DNNs are explicitly constrained to process words in an order dictated by a parse tree, rather than from left to right, their syntactic behavior more closely matches that of humans: they require less data than standard DNNs to acquire certain generalizations, and they generalize in a human-like way to new syntactic structures \citep[see also][]{Hale:etal:2018,Kuncoro:etal:2018a}. At this point in time, the technological tools for injecting linguistic constraints into standard DNN architectures are still in their infancy; we do not yet have reliable methods to implement proposals for innate constraints that are more specific than a general sensitivity to the parse of a sentence. Developing such tools would significantly benefit linguistics and cognitive science, and, we argue, is an important area for future research. In the meantime, negative results must be taken with more than one grain of salt, as they might reflect the technological difficulty of combining neural networks and symbolic knowledge, rather than the inefficacy of the linguistic priors in question \emph{per se}.
\\
\\
\noindent{}\textbf{Amount and nature of training data.} 
Neural networks extract linguistic generalizations from raw, unannotated language data, or so the standard spiel goes. But to understand the implications of the successes and failures of a particular DNN experiment, we need to consider the nature of the training data used in the experiment. A first fundamental division is between \emph{supervised/focused} and \emph{unsupervised/generic} training. Consider for example the difference between the studies of \citet{Linzen:etal:2016} and \citet{Gulordava:etal:2018}, both of which suggest that DNNs can learn (long-distance) agreement (Section~\ref{sec:long-distance-agreement}). In Linzen's supervised setup, the network was fed many sentences exemplifying subject-verb agreement, and was explicitly trained on the objective of predicting verb number. This study thus asked the following question: Is the architecture of a DNN able, in principle, to learn agreement, even if explicit instruction is required?  By contrast, Gulordava and colleagues took a network that was trained to predict each word in a corpus from its context, and investigated, with no further instruction, whether it displayed sensitivity to agreement dependencies. The question here changes to: Is there enough signal in a raw text corpus for the DNN architecture in question to correctly pick up the agreement rule? 

Another central distinction is between \emph{synthetic} and \emph{corpus-extracted} test (or training) data. The agreement benchmarks of Linzen and Gulordava were derived from corpora; other studies trained the DNN on corpus data but tested it on constructed examples \citep[e.g.,][]{Marvin:Linzen:2018,Futrell:etal:2019}, or trained \textit{and} tested the DNN on synthetic data \citep[e.g.,][]{McCoy:etal:2020}. All of these setups may lead to useful insight, but the interpretation of the results should change accordingly, and different confounds have to be taken into account. For example, unfiltered corpus data may contain spurious correlations that the network could rely upon \citep{Kuncoro:etal:2018b}. On the other hand, synthetic sentences might be so different from the corpus data the DNN was trained on that its failure to handle them might have more to do with irrelevant differences in factors such as lexical frequency distributions than with the grammatical phenomenon under investigation.

Even when the training data consist of a raw corpus of naturally occurring language, as they do in popular NLP word prediction models such as BERT \citep{Devlin:etal:2019} and GPT-2 \citep{Radford:etal:2019}, it is important to remember that these data differ significantly from those that a child is exposed to, in both size and nature. In NLP, the old adage that ``more data is better data'' \citep{Banko:Brill:2001} has held up remarkably. Over the course of just one year, NLP practitioners have increased the size of the corpora used to train word prediction models from 4 billion words (BERT) to 8 billion words (GPT-2) to well over 100 billion words (T5; \citealt{raffel2019exploring}). Clearly, the steadily increasing amount of data made available to these systems is orders-of-magnitude larger than that available to children (at most 10 million a year, according to \citealt{hart1995meaningful}). This divergence limits the cognitive conclusions that can be drawn from testing off-the-shelf NLP systems, and implies that linguists need to train their own DNNs to assess how the amount of training data impacts DNNs' syntactic knowledge \citep{vanschijndel2019quantity}.  
DNNs' training data differ from those of children not only in size but also in nature. The books and articles that DNNs are trained on in NLP differ in their syntactic properties from child-directed speech. More generally, word prediction networks in NLP are, in essence, asked to learn language by reading entire libraries of books while sitting alone in a dark room. This setting is profoundly different from the context in which children learn language, which is grounded in perception and social interaction. Efforts are underway to introduce such grounding into DNNs---see, for example, \cite{Chrupala:etal:2015} and \cite{Weston:2016}, respectively---but these efforts are severely limited by the paucity of appropriate training data. When making claims about the syntactic abilities that can be acquired by the DNNs discussed in this survey, then, we must keep in mind that the real question we are asking is how much can be learned from huge amounts of written linguistic data alone. While purely distributional cues are one of the sources of information used by children when acquiring syntax \citep{Gomez:Gerken:2000}, they are certainly not the only type of evidence children rely upon; this greatly complicated the quantitative comparison between the amount of data available to children and DNNs.
\\
\\
\noindent{}\textbf{Implications for the study of human linguistic abilities.}
The body of work we reviewed in this survey establishes that DNNs are capable of high accuracy in challenging syntactic tasks such as implicitly distinguishing the head noun of the subject from other nouns in the sentence. At the same time, nearly all studies reported that DNNs' behavior deviated from the idealized syntactic competence that a linguist might postulate; in the case of agreement, for instance, their behavior suggests that they rely on a complex array of heuristics, rather than on a fully-fledged context-free grammar that would allow them to correctly process center embedded clauses, such as Example~\ref{multiple-embeddings} above. In what ways, then, can findings on the syntactic abilities of DNNs inform the cognitive science of language in humans?

Minimally, DNNs can be useful to thoroughly vet the stimuli of human language processing experiments. If a DNN that is known to fall short of human competence succeeds on a task, this suggests that the task in question may not probe the full-fledged grammatical abilities we think humans possess. Consider for example the experiment of \citet{Gulordava:etal:2018}, who showed that DNNs are almost as good as Italian speakers in the long-distance agreement task. This suggests that the stimuli that were used did not truly probe the human ability to resort to a full-fledged context-free grammar to parse sentences, or else the difference between DNN and human subjects would have been much more dramatic, given the limitations uncovered by \citet{Lakretz:etal:2019} and \citet{Marvin:Linzen:2018}.

At the same time, humans themselves often deviate from linguists' idealized description of syntactic \emph{competence}. We regularly make agreement errors \citep{Bock:Miller:1991}, and have difficulty parsing multiply center-embedded sentences \citep{chomsky1963miller,gibson1999memory}. DNN error patterns, and the heuristics that give rise to those errors, may therefore serve as a source of hypotheses for experiments designed to study human syntactic \emph{performance} \citep{Linzen:Leonard:2018}. 

Going even further, not all modern linguistic theorists recognize a sharp distinction between competence and performance. Under some views of syntax, grammar is more akin to a ``toolbox of tricks'' we picked up along our evolutionary way than to the maximally elegant and powerful formal grammars of computer science \citep[e.g.,][]{Culicover:Jackendoff:2005,Pinker:Jackendoff:2005,Goldberg:2019}. Under such views, the difference between the syntactic knowledge of DNNs and that of humans might be more one of quantity than quality: humans possess a larger and more sophisticated set of heuristics to parse sentences than DNNs do, but they do not rely on any radically different and more powerful ``narrow language faculty'' abilities. If that is the case, the behavior of DNNs might give us insights not only into online processing (``performance''), but also on some of the core syntactic tools that constitute human grammatical competence. We look forward to theoretical work linking modern DNNs to construction grammar and similarly ``shallow'' syntactic formalisms.
\\
\\
\noindent{}\textbf{Conclusion.} In our view, the time is ripe for more linguists to get engaged in the lines of work we sketched in this survey. On the one hand, linguists' know-how in probing grammatical knowledge can help develop the next generation of language-processing DNNs, and the success of events such as the BlackBox NLP series confirms that the deep learning community is warmly welcoming linguistic analyses of DNNs. On the other, studying what the best DNNs learn about grammar, and how they do so, can offer new insights about the nature of language and, ultimately, what is genuinely unique about the human species. For this line of work to be effective, linguists will need to be closely involved in developing relevant network architectures, training them on appropriate data, and conducting experiments that address linguists' theoretical concerns.

\section*{ACKNOWLEDGMENTS}
Tal Linzen was supported by a Google Faculty Research Award, National Science Foundation grant BCS-1920924, and the United States--Israel Binational Science Foundation (award 2018284).
We thank Robert Frank, Ethan Wilcox and the JHU Computation and Psycholinguistics Lab for discussion of the issues reviewed in the article.

\bibliography{structure_from_deep_learning}
\bibliographystyle{ar-style1}

\end{document}